\documentclass[11pt, twocolumn]{article}

\usepackage[margin=0.75in]{geometry}
\usepackage{graphicx}
\usepackage{amsmath, amssymb}
\usepackage{booktabs}
\usepackage{hyperref}
\usepackage{enumitem}
\usepackage{xcolor}
\usepackage{float}
\usepackage{caption}
\usepackage{subcaption}
\usepackage{titlesec}
\usepackage{tikz}
\usepackage{mdframed}
\usetikzlibrary{positioning, arrows.meta, fit, backgrounds}

\graphicspath{{./}}

\titlespacing*{\section}{0pt}{10pt}{5pt}
\titlespacing*{\subsection}{0pt}{8pt}{4pt}
\titlespacing*{\subsubsection}{0pt}{6pt}{3pt}

\title{
\textbf{Passive Construction Site Safety Monitoring via Persona-Scaffolded Adversarial Chain-of-Thought VLM Verification} \\[6pt]
}

\author{
Ananth Sriram$^*$ \quad Neel Mokaria$^*$ \quad Rajveer Singh$^*$ \\[4pt]
\small $^*$Department of Computer Science, University of Maryland, College Park, MD, USA \\
\small \texttt{\{asriram2, nmokaria, rajveer\}@umd.edu} \\[2pt]
\small $^\dagger$Authors listed in alphabetical order; all contributed equally.
}

\date{February 2026}

\begin{document}
\maketitle

\begin{abstract}
Construction remains the deadliest industry sector in the United States,
with 1,055 fatal worker injuries recorded in 2023 at a rate of 12.9 deaths
per 100,000 full-time equivalent workers. The majority of these incidents
are preventable, yet existing monitoring approaches are either
prohibitively expensive, require real-time human operators, or address
only a narrow subset of detectable violations. This paper presents a passive, end-of-shift
construction safety monitoring pipeline, processing video
from two independent camera modalities --- POV body-worn and fixed
wall-mounted cameras --- through a unified three-stage architecture:
(1) a fine-tuned YOLO11 detector for primary PPE and hazard detection,
(2) SAM~3 for segmentation refinement, worker deduplication, and
PPE-to-worker association, and (3) Qwen3-VL-8B-Instruct with a novel
three-pass adversarial chain-of-thought protocol for compliance
verification and hallucination control. The
principal methodological contribution is the Stage~3 prompt design: we
find that the specific construction of professional persona backstories,
following the method-actor prompt engineering framing, drives an observed 
12\% precision improvement on an informal three-author review of the 12-video Ironsite video corpus, with the largest
gains on hallucination-prone categories. Structural message isolation
enforces observational independence between a generator pass, a
discriminator pass, and a reconciliation pass governed by asymmetric
rules that encode explicit priors about the relative reliability of human
observation versus automated detection for different violation categories.
The system maps detected violations to specific OSHA standards, performs
REBA-inspired ergonomic risk scoring from pose keypoints, and produces
per-worker safety reports with timestamped frame-level evidence. A
quantitative evaluation harness is released alongside the codebase for
future reproduction.
\end{abstract}

\section{Introduction}

The Bureau of Labor Statistics reported a fatal work injury rate of 12.9 deaths per 100,000 full-time equivalent workers in the U.S. construction sector in 2023, making it the industry with the most fatalities of any sector~\cite{bls2023}. Men accounted for 91.5\% of fatalities. Fall protection (OSHA Standard 1926.501) alone accounts for the single largest violation category cited by OSHA year over year~\cite{osha2024}.

The operational gap we target is the lack of an affordable, passive, continuous monitoring approach that can watch multiple workers across a construction site, resolve their identities over time, and produce an actionable report without requiring real-time human review. Existing solutions tend to be expensive, require dedicated real-time operators, or cover only a narrow slice of the problem---for instance, hard hat detection alone.

We explore an \textit{end-of-shift} processing paradigm as one possible response. Rather than streaming video in real time, the system processes footage after a shift completes, which makes it more practical to deploy on commodity hardware while using higher-capacity models (including an 8-billion-parameter vision-language model) and multi-pass verification that would be difficult to support under real-time latency constraints.

The contributions of this paper are descriptive and architectural:
\begin{enumerate}[nosep, leftmargin=*]
    \item A unified three-stage pipeline (YOLO $\rightarrow$ SAM~3 $\rightarrow$ VLM) that handles both PPE and posture/ergonomic analysis from two independent camera modalities.
    \item A method-prompted, persona-scaffolded three-pass adversarial chain-of-thought VLM protocol --- following the method-actor framing of Doyle~\cite{doyle2024methodactors} --- with structurally enforced independence between passes and an asymmetric reconciliation policy, which yields an observed 12\% precision improvement over single-pass prompting on the development corpus.
    \item REBA-inspired ergonomic risk scoring from YOLO pose keypoints with tuned confidence gates.
    \item An end-of-shift backend that maps detected events to specific OSHA standards and produces per-worker safety reports with timestamped evidence.
\end{enumerate}

\section{Related Work}

\subsection{PPE Detection in Construction Environments}

Automated detection of personal protective equipment (PPE) using computer 
vision has been an active area of research since the widespread adoption of 
deep convolutional neural networks. Early work by Fang et al.~\cite{fang2018} 
demonstrated that Faster R-CNN could detect hard hat compliance from far-field 
surveillance footage with sufficient accuracy for practical deployment, 
establishing the feasibility of passive vision-based monitoring in outdoor 
construction environments. Nath et al.~\cite{nath2020} extended this line of 
work to multi-class PPE detection using YOLOv3, reporting mean average 
precision (mAP) values above 0.85 on construction-specific datasets. More 
recent studies have leveraged YOLOv5 and YOLOv8 architectures on larger 
composite datasets, consistently demonstrating that detection accuracy 
degrades under variable illumination, occlusion, and worker 
clustering~\cite{Sanjeewanietal2024}. A persistent limitation across this 
body of work is the restriction to a single PPE category or a single camera 
viewpoint, without integration of ergonomic risk assessment or persistent 
worker identity tracking across a shift.

\subsection{Ergonomic Risk Assessment via Pose Estimation}

Traditional ergonomic risk assessment in construction relies on manual 
observation instruments, principally the Rapid Entire Body Assessment 
(REBA)~\cite{hignett2000} and Rapid Upper Limb Assessment (RULA), which 
require trained observers and are inherently episodic rather than continuous. 
Computer vision approaches to automating these assessments have drawn 
primarily on skeleton-based pose estimation. Yan et al.~\cite{yan2017} 
proposed wearable IMU-based systems to detect musculoskeletal disorder (MSD) 
risk in real time, achieving high sensitivity for trunk flexion and arm 
overreach events. Subsequent work demonstrated that markerless pose estimation 
using convolutional neural network-based keypoint models can recover joint 
angles sufficient for ergonomic scoring from monocular video without 
instrumented garments~\cite{Zhaoetal2024}. However, these systems have been 
validated primarily in controlled laboratory settings; their performance on 
outdoor construction footage with partial occlusion and variable depth remains 
incompletely characterized.

\subsection{Worker Re-identification and Identity Persistence}

Maintaining persistent worker identities across a full shift is a prerequisite 
for accumulating violation histories and generating per-worker compliance 
reports, yet it has received comparatively limited attention in the 
construction safety literature. Detection-based tracking systems operating 
in construction environments must contend with visual confounds introduced 
by high-visibility PPE --- similar vest colours and helmet profiles reduce 
the discriminative power of appearance-based 
features~\cite{Gengetal2025}. BoT-SORT~\cite{Aharonetal2022} and related 
tracking-by-detection algorithms address short-term occlusion effectively 
through coupled Kalman filter motion prediction and appearance 
re-identification, but are not designed for the multi-hour gaps characteristic 
of shift-level tracking. The present work adopts a lightweight identity 
database combining BoT-SORT for within-session tracking with appearance 
embedding similarity for re-association after disappearance events, 
representing a pragmatic intermediate between pure tracking and full Re-ID.

\subsection{Vision-Language Models for Safety Assessment}

The application of large vision-language models (VLMs) to safety-critical 
inspection tasks is nascent but growing. Generalist VLMs have 
shown promising zero-shot performance on visual question answering 
tasks involving workplace safety scenes, yet their deployment in 
safety-critical contexts is complicated by hallucination --- the generation 
of plausible but factually unsupported outputs~\cite{Baietal2024}. 
Chain-of-thought prompting, while improving task accuracy on reasoning 
benchmarks, has been reported to induce systematic overconfidence in VLMs by 
causing token probabilities to reflect consistency with the model's own 
reasoning trace rather than uncertainty about visual 
evidence~\cite{Welchetal2025}. Multi-agent and self-consistency prompting 
approaches~\cite{wang2022} have demonstrated that soliciting multiple 
independent completions and aggregating by consistency can reduce hallucination 
rates, which motivates the multi-pass verification protocol described 
in this work. To the best of our knowledge, prior published work has not 
applied this style of multi-pass adversarial VLM verification specifically 
to construction site PPE or ergonomic compliance assessment.

\subsection{Integrated Construction Safety Monitoring Systems}

Several research systems have proposed integrating multiple safety sensing 
modalities for construction site monitoring. Kim et al.~\cite{Kimetal2016} 
proposed a real-time system combining RFID-based proximity sensing with 
camera-based PPE detection for zone-level compliance monitoring, but required 
dedicated infrastructure installation. Awolusi et al.~\cite{Awolusietal2018} 
demonstrated wearable sensor fusion for ergonomic and fall risk assessment 
but noted that per-worker hardware acquisition and maintenance costs represent 
a significant barrier to large-site deployment. A distinguishing 
characteristic of the present system is its \textit{end-of-shift} processing 
paradigm, which decouples monitoring quality from real-time compute 
constraints. This makes it feasible to use large segmentation and 
vision-language models that would be difficult to run under streaming 
latency requirements, while relying only on standard video capture 
infrastructure already present on many modern construction sites.
\section{Methodology}

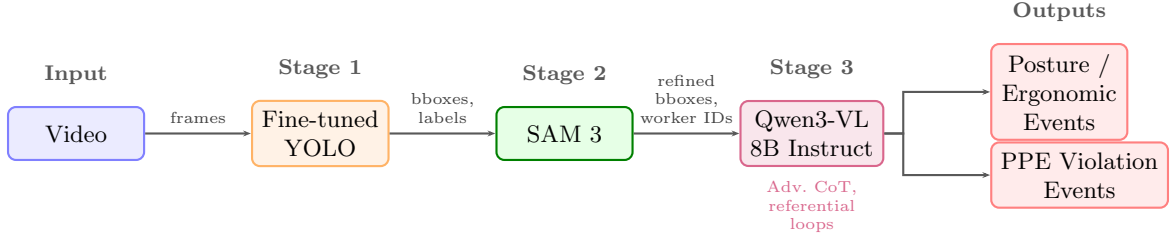
\begin{figure*}[t]
\centering
\begin{tikzpicture}[
    node distance=0.4cm and 0.8cm,
    >={Stealth[length=3pt]},
    stage/.style={draw, rounded corners=3pt, minimum height=0.7cm, minimum width=1.8cm, font=\footnotesize, align=center, thick},
    input/.style={stage, fill=blue!8, draw=blue!50},
    s1/.style={stage, fill=orange!10, draw=orange!60},
    s2/.style={stage, fill=green!10, draw=green!50!black},
    s3/.style={stage, fill=purple!10, draw=purple!60},
    outstyle/.style={stage, fill=red!8, draw=red!50},
    grplabel/.style={font=\scriptsize\bfseries, text=gray!70!black},
    arr/.style={->, thick, gray!70!black},
    arrlabel/.style={font=\tiny, text=gray!50!black, midway},
]
\node[input] (vid) {Video};

\node[s1, right=1.4cm of vid] (yolo) {Fine-tuned\\YOLO};

\node[s2, right=1.4cm of yolo] (sam) {SAM~3};

\node[s3, right=1.4cm of sam] (vlm) {Qwen3-VL\\8B Instruct};

\node[outstyle, right=1.4cm of vlm, yshift=0.55cm] (pos) {Posture /\\Ergonomic\\Events};
\node[outstyle, right=1.4cm of vlm, yshift=-0.55cm] (ppe) {PPE Violation\\Events};

\node[grplabel, above=0.15cm of vid] {Input};
\node[grplabel, above=0.15cm of yolo] {Stage 1};
\node[grplabel, above=0.15cm of sam] {Stage 2};
\node[grplabel, above=0.15cm of vlm] {Stage 3};
\node[grplabel, above=0.15cm of pos] {Outputs};

\draw[arr] (vid) -- node[arrlabel, above] {frames} (yolo);
\draw[arr] (yolo) -- node[arrlabel, above, text width=1.2cm, align=center] {bboxes,\\labels} (sam);
\draw[arr] (sam) -- node[arrlabel, above, text width=1.4cm, align=center] {refined bboxes,\\worker IDs} (vlm);
\draw[arr] (vlm.east) -- ++(0.25,0) |- (pos.west);
\draw[arr] (vlm.east) -- ++(0.25,0) |- (ppe.west);

\node[font=\tiny, text=purple!60, below=0.05cm of vlm, text width=1.8cm, align=center] {Adv.\ CoT,\\referential loops};

\end{tikzpicture}
\caption{Three-stage pipeline architecture shared by both posture/ergonomics and PPE violation detection. Video frames pass through YOLO detection, SAM~3 refinement, and adversarial VLM verification to produce compliance-checked safety events.}
\label{fig:pipeline}
\end{figure*}

The proposed system processes construction site video through a unified three-stage pipeline
shared across both the posture/ergonomics and PPE violation detection streams.
Artifacts propagate sequentially: Stage~1 produces initial detections, Stage~2 refines
spatial evidence and resolves worker identity, and Stage~3 applies a vision-language model
to assess compliance against OSHA standards with measures intended to reduce hallucination.
Figure~\ref{fig:pipeline} illustrates the end-to-end data flow, and
Table~\ref{tab:pipeline} summarizes each stage's role and outputs. The principal methodological contribution is the prompt design for Stage~3: a method-prompted, persona-scaffolded adversarial chain-of-thought protocol
that produced an observed 12\% precision improvement on a dev-team review over single-pass VLM
prompting on the Ironsite development corpus (Section~\ref{sec:results}).

\begin{table*}[t]
\centering
\small
\caption{Three-stage pipeline: components, roles, and outputs.}
\label{tab:pipeline}
\begin{tabular}{cl p{5.2cm} p{4.8cm}}
\toprule
\textbf{Stage} & \textbf{Component} & \textbf{Role} & \textbf{Output} \\
\midrule
1 & Fine-tuned YOLO & First pass on video: detect workers, PPE items, poor-posture candidates, violation candidates & Initial bounding boxes, class labels, frame-level detections \\[3pt]
2 & SAM~3 & Worker tagging, deduplication, refined masks and bboxes (MoE segmentation) & Refined bboxes, worker/track IDs, segmentation masks \\[3pt]
3 & Qwen3-VL-8B-Instruct & Assesses Stage 1--2 outputs via method prompting, chain-of-thought, and referential loops; applies OSHA mapping; targets hallucination reduction & Compliance-checked violation flags $\rightarrow$ final posture/ergonomic and PPE violation outputs \\
\bottomrule
\end{tabular}
\end{table*}

\subsection{Stage 1: Primary Detection}

A YOLO11 object detector was fine-tuned on a composite dataset assembled from
five construction-specific PPE sources: the Ultralytics Construction-PPE collection
(11 classes), the Roboflow construction site safety dataset, the SH17 dataset,
and two additional PPE corpora sourced from Kaggle. The resulting detector recognizes
12 mapped semantic classes spanning worker presence, PPE compliance state, and
proximate equipment, including hard hats, safety vests, gloves, and proximate
machinery such as excavators, cranes, and forklifts.

Detection confidence is set deliberately low at $\tau_{\text{conf}} = 0.15$ with the
intent of maximizing recall at this stage; the downstream VLM verification pass is
responsible for suppressing false positives. This design reflects a principled asymmetry:
in a safety monitoring context, the cost of a missed violation is typically considered to
exceed the cost of a spurious detection that is subsequently rejected.

For the wall-mounted camera stream, YOLO operates in conjunction with BoT-SORT~\cite{Aharonetal2022},
a multi-object tracker that maintains persistent track identifiers across frames via
coupled Kalman filter motion prediction and appearance re-identification. This enables
temporal PPE accumulation per tracked worker: rather than issuing a violation on a
single frame, the system evaluates compliance over a sliding window of detections,
which is intended to reduce the impact of transient occlusions and detector noise.

For the body-worn camera stream, a separate YOLO pose estimation model extracts
17 COCO-format skeletal keypoints per detected person per frame, providing the
joint coordinate sequences required for downstream ergonomic angle computation.

\subsection{Stage 2: Spatial Refinement and Worker Deduplication}

Bounding boxes produced by Stage~1 are forwarded to SAM~3 (Segment Anything
Model~3)~\cite{ravi2024sam2} as prompted inputs to generate pixel-accurate segmentation masks.
This stage serves three functions within the pipeline.

First, SAM~3 masks are used to tighten the bounding boxes produced by YOLO, correcting for
the tendency of detection models to over-expand boxes in the presence of adjacent
workers or equipment. Tighter crops are intended to improve both the visual evidence
quality presented to the VLM and the accuracy of spatial proximity computations used
for hazard detection.

Second, per-worker segmentation masks provide more semantically isolated inputs for
VLM processing. Rather than presenting the full scene to the language model,
Stage~2 produces isolated worker crops with overlaid annotation, which we use to
reduce the contextual ambiguity that has been identified as a source of hallucination
in VLM-based assessment~\cite{Baietal2024}.

Third, PPE-to-worker association is resolved via center-point containment:
for each detected PPE item, the system determines which worker's bounding box
contains the centroid of the PPE mask. This heuristic addresses the common
failure mode of assigning a detected hard hat to an adjacent rather than the
proximate worker when workers are clustered. Required PPE per worker is
defined as $\mathcal{P}_{\text{req}} = \{\text{hardhat}, \text{safety\_vest},
\text{gloves}\}$; a violation is flagged if any element of $\mathcal{P}_{\text{req}}$
is unobserved after a temporal accumulation window.

\subsection{Stage 3: Compliance Verification via Adversarial Vision-Language Model Assessment}

The third stage applies a fine-tuned Qwen3-VL-8B-Instruct model
augmented with an optional LoRA adapter for construction-domain specialization. The central methodological contribution of this work lies in the Stage~3 prompt design. We find that the specific construction of persona backstories
--- not merely the use of personas, but the granularity of role,
inspection mandate, and evidence access scoped to each persona --- has
substantial impact on output precision. On the Ironsite development corpus,
the full three-pass adversarial protocol introduced below produced an observed 12\%
precision improvement over single-pass prompting of the same Qwen3-VL-8B
base model with an equivalent token budget. The central design challenge at this stage is hallucination: large VLMs prompted on safety-critical visual evidence have been reported to exhibit overconfidence,
generating plausible but factually unsupported violation reports~\cite{Baietal2024}.
Single-pass prompting is particularly susceptible because the model has no mechanism
for cross-checking its own outputs against independent evidence.

To address this, a \textbf{three-pass adversarial chain-of-thought} protocol is introduced,
drawing on the self-consistency principle~\cite{wang2022} and extending it toward
multi-perspective inspection rather than simple output aggregation.

\textbf{Pass~1 (Generation):} A field safety inspector persona reviews the raw,
unannotated video stream in isolation, with no access to machine detection outputs.
This pass produces free-form inspection notes grounded solely in visual observation,
with the intent of establishing an independent evidence signal that is not anchored
to Stage~1--2 detector artifacts.

\textbf{Pass~2 (Discrimination):} A senior safety officer persona independently
reviews the same video alongside YOLO/SAM~3 annotated frames---bounding boxes,
segmentation masks, and confidence scores overlaid. This pass has no access to
Pass~1 notes and is tasked with validating or rejecting each Stage~1--2 machine
detection from an independent observational perspective.

\textbf{Pass~3 (Reconciliation):} The same senior persona reconciles all available
evidence---the Pass~1 observational notes, the Pass~2 machine-detection assessment,
and the raw YOLO/SAM~3 confidence data---into a structured \texttt{VLMAssessment}
object. Conflicts between Pass~1 and Pass~2 assessments are resolved through
explicit chain-of-thought reasoning, with the reconciler required to justify any
departure from either upstream signal.

The adversarial structure is intended to reduce hallucination through two complementary
mechanisms. First, cross-pass consistency checking: a violation reported in Pass~3
should be traceable to at least one of the two independent upstream signals,
discouraging the model from generating violations supported by neither visual observation
nor machine detection. Second, confidence calibration: YOLO detection confidence scores
are explicitly communicated to the VLM as prior signal strength, instructing the
model to treat high-confidence detections ($\geq 0.70$) as strong priors requiring
rebuttal evidence to override, and low-confidence detections ($< 0.40$) as weak
signals requiring independent visual confirmation before violation issuance
(Table~\ref{tab:conf-bands}).

Long-form videos are segmented into 60-second non-overlapping chunks prior to
VLM processing to prevent context window overflow; each chunk receives an independent
assessment. The final output per chunk is a structured \texttt{VLMAssessment}
containing: a scene summary, worker count, equipment inventory, per-worker PPE
compliance status (each item classified as \texttt{PRESENT}, \texttt{ABSENT}, or
\texttt{UNCLEAR}), a typed hazard list with severity classification and explanatory
rationale, aggregate confidence, and actionable corrective recommendations.

\subsection{Prompt Design and Adversarial Independence}
\label{sec:prompt-design}

The three-pass protocol is not merely a stylistic prompt variation: persona separation
is enforced at the message-construction level. Pass~1's message array contains no
YOLO detection data and no annotated frames; Pass~2's message array contains no
Pass~1 output. Independence is therefore structural rather than instructional ---
a model operating in Pass~2 does not have Pass~1 observations in its context,
which is intended to reduce the cross-contamination that would arise if both
passes shared a common context window.

Table~\ref{tab:pass-inputs} summarises the input modalities available to each pass.
The full system prompts for each pass are provided in the Supplementary Material
(Appendix~\ref{app:prompts}).

\begin{table}[H]
\centering
\small
\caption{Input modalities per VLM pass.}
\label{tab:pass-inputs}
\resizebox{\columnwidth}{!}{
\begin{tabular}{lcccc}
\toprule
\textbf{Pass} & \textbf{Video} & \textbf{Frames} & \textbf{YOLO} & \textbf{Notes} \\
\midrule
1 (Generate)     & \checkmark & --- & --- & --- \\
2 (Discriminate) & \checkmark & \checkmark & \checkmark & --- \\
3 (Reconcile)    & ---        & \checkmark & \checkmark & Both \\
\bottomrule
\end{tabular}
}
\end{table}

Pass~3 deliberately omits the raw video. This is a forcing function: the reconciler
operates on \textit{committed evidence} --- the written observations each pass has
already committed to in natural language --- rather than re-litigating ambiguous
video frames. The structured \texttt{VLMAssessment} output (Table~\ref{tab:schema})
is produced exclusively by Pass~3, which must justify each hazard decision against
all three upstream evidence sources.

\begin{table}[H]
\centering
\small
\caption{VLMAssessment output schema (Pydantic-validated).}
\label{tab:schema}
\resizebox{\columnwidth}{!}{
\begin{tabular}{ll}
\toprule
\textbf{Field} & \textbf{Type / Values} \\
\midrule
\texttt{scene\_summary}     & string \\
\texttt{worker\_count}      & int $\geq 0$ \\
\texttt{equipment\_present} & list[string] \\
\texttt{ppe\_per\_worker}   & list[PPEStatus] \\
\texttt{reasoning}          & string (mandatory audit trail) \\
\texttt{hazards}            & list[HazardDetail] \\
\texttt{no\_hazards}        & bool \\
\texttt{confidence}         & LOW $|$ MEDIUM $|$ HIGH \\
\midrule
\multicolumn{2}{l}{\textit{PPEStatus fields:}} \\
\texttt{helmet, vest, gloves} & PRESENT $|$ ABSENT $|$ UNCLEAR \\
\midrule
\multicolumn{2}{l}{\textit{HazardDetail fields:}} \\
\texttt{violation\_type}    & PPE\_MISSING $|$ FALL\_PROTECTION\_MISSING $|$ \\
                            & PROXIMITY\_HAZARD $|$ ZONE\_BREACH $|$ \\
                            & LADDER\_MISUSE $|$ SCAFFOLD\_VIOLATION $|$ \\
                            & BEHAVIORAL\_UNSAFE $|$ AWKWARD\_POSTURE $|$ \\
                            & MSD\_HIGH\_RISK $|$ OVERREACH $|$ \\
                            & KNEELING\_SQUATTING\_LOW \\
\texttt{severity}           & LOW $|$ MEDIUM $|$ HIGH $|$ CRITICAL \\
\texttt{best\_frame\_index} & int (0-based index into annotated frames) \\
\bottomrule
\end{tabular}
}
\end{table}

The \texttt{best\_frame\_index} field grounds each hazard to a specific annotated
frame by interleaving \texttt{[Annotated frame i]} text markers before each image
in the Pass~3 message array, enabling the post-processor to resolve the index to
a filename for timestamped evidence retrieval.

\subsubsection{YOLO Confidence Calibration}

Passes~2 and~3 receive YOLO detection data rendered as structured text, with
per-class confidence scores formatted as:

\begin{small}
\begin{verbatim}
t=12.40s | detections: person=0.91
           helmet=0.42  vest=0.18
  hazards: PPE_MISSING (worker #3)
           [helmet=0.42, vest=0.18]

Worker #3 [VIOLATION]
  present=[helmet]  missing=[vest, gloves]
\end{verbatim}
\end{small}

A four-band calibration scale is injected into the Pass~2 and Pass~3 prompts,
explicitly mapping confidence values to epistemic weight:

\begin{table}[H]
\centering
\small
\caption{YOLO confidence calibration bands injected into Passes 2 and 3.}
\label{tab:conf-bands}
\resizebox{\columnwidth}{!}{
\begin{tabular}{lll}
\toprule
\textbf{Confidence} & \textbf{Signal} & \textbf{VLM instruction} \\
\midrule
$\geq 0.70$         & HIGH     & Strong prior: detection is real \\
$0.40$--$0.69$      & MODERATE & Verify visually \\
$0.15$--$0.39$      & WEAK     & Must confirm in video \\
$< 0.15$            & NOISE    & Discard unless confirmed \\
\bottomrule
\end{tabular}
}
\end{table}

Pass~1 receives no confidence data by design: the generator persona is intended
to provide an observational baseline that is not anchored to machine output.

\subsubsection{Asymmetric Reconciliation Rules}

A key element of the prompt design is a set of asymmetric reconciliation
rules governing how Pass~3 resolves disagreements between the two observers and
the machine detection signal. These rules encode an explicit prior that reflects
the complementary strengths and weaknesses of human observation versus automated
detection in construction safety contexts:

\begin{enumerate}[nosep, leftmargin=*]
    \item \textbf{Agreement} (both observers and YOLO concur): flag with high
    confidence.
    \item \textbf{Generator-only flag} (Pass~1 flagged, Pass~2 did not): do not
    automatically dismiss. Pass~1 observed raw video and is sensitive to motion,
    behaviour, and transient events --- phone use, running, brief PPE removal ---
    that annotated still frames may not capture. A specific reason to overrule is
    required; absence of machine detection is insufficient.
    \item \textbf{Discriminator-only flag} (Pass~2 flagged, Pass~1 did not):
    flag if machine evidence is strong ($\text{conf} \geq 0.60$).
    \item \textbf{Machine-only detection} (neither observer flagged): flag with
    explanatory note if $\text{conf} \geq 0.70$; discard if $\text{conf} < 0.40$.
    \item \textbf{Observer-only flag} (either observer flagged, YOLO missed):
    flag. Automated detectors are known to miss behavioural
    violations, and a clear human observation is treated here as sufficient
    evidence.
    \item \textbf{Conflict} (observers disagree): rule in favour of the observer
    with clearer contextual access to that violation type --- Pass~1 for
    behavioural and motion violations; Pass~2 for spatial and PPE violations
    corroborated by machine detection.
\end{enumerate}

These rules instantiate an intentional asymmetry: behavioural violations
(phone use, unsafe movement, zone breaches) are treated as
\textit{observer-strong, machine-weak}, while spatial and PPE violations
are treated as \textit{machine-strong, observer-weak}. This asymmetry
is motivated by our qualitative observations of the detection
characteristics of each modality on construction footage, and is intended
to reduce the false negative rate for the violation categories most
responsible for construction fatalities~\cite{osha2024}. The correctness
and calibration of this asymmetry remain to be validated quantitatively.

\subsubsection{System Prompts}

The full system prompts used for Passes 1--3 are reproduced verbatim in the
Supplementary Material (Appendix~\ref{app:prompts}) to support reproducibility.
Camera-modality-specific addenda (injected for POV and wall-mounted variants)
are documented in the released codebase.

\section{Posture and Ergonomic Analysis}

For POV body-worn camera footage, ergonomic risk is assessed via a REBA-inspired
pipeline operating on skeletal keypoints extracted by the YOLO pose model.

\subsection{Joint Angle Extraction}

Six angles are computed from the 17 COCO keypoints per frame: trunk flexion
($\theta_{\text{trunk}}$), trunk lateral lean ($\theta_{\text{lat}}$), neck
flexion ($\theta_{\text{neck}}$), knee angle ($\theta_{\text{knee}}$), arm
raise ($\theta_{\text{arm}}$), and elbow flexion ($\theta_{\text{elbow}}$).
A keypoint confidence gate of $\tau_{\text{kp}} = 0.65$ is enforced to
suppress detections from occluded or partially visible poses.

\subsection{Risk Scoring and Violation Classification}

Angles are combined into a scalar risk score following the two-group REBA
scheme~\cite{hignett2000}:

\begin{align}
\text{Score}_A &= f(\theta_{\text{trunk}}, \theta_{\text{neck}},
\theta_{\text{knee}}, \theta_{\text{lat}}) \\
\text{Score}_B &= g(\theta_{\text{arm}}, \theta_{\text{elbow}}) \\
\text{Combined} &= \max(\text{Score}_A, \text{Score}_B) + \Delta
\end{align}

where $\Delta$ captures joint interaction effects. The combined score maps
to four violation levels (Table~\ref{tab:posture}), with specific triggers
including arm elevation $> 65^\circ$ with concurrent body twist
(\texttt{OVERREACH}), trunk flexion $> 48^\circ$ (\texttt{AWKWARD\_POSTURE}),
and trunk flexion $> 65^\circ$ (\texttt{MSD\_HIGH\_RISK}).

\begin{table}[H]
\centering
\small
\caption{Posture violation classification thresholds.}
\label{tab:posture}
\resizebox{\columnwidth}{!}{
\begin{tabular}{lll}
\toprule
\textbf{Condition} & \textbf{Type} & \textbf{Risk} \\
\midrule
Combined $< 3$    & Compliant              & Level 1 \\
Combined $\geq 3$ & \texttt{AWKWARD\_POSTURE} & Level 2 \\
Combined $\geq 5$ & \texttt{OVERREACH}        & Level 3 \\
Combined $\geq 8$ & \texttt{MSD\_HIGH\_RISK}  & Level 4 \\
\bottomrule
\end{tabular}
}
\end{table}

\section{Worker Identity and Temporal Accumulation}

Worker identity is maintained across a shift via two complementary mechanisms.
BoT-SORT~\cite{Aharonetal2022} provides frame-to-frame track continuity through
Kalman filter motion prediction, handling short-term occlusions. For longer
disappearances, a lightweight identity database stores per-worker appearance
embeddings and re-associates returning workers by embedding similarity,
which is intended to prevent duplicate identity assignment.

PPE compliance is evaluated over a temporal accumulation window rather than
on individual frames. Per-worker logs track observed PPE items across the chunk;
a violation is issued only if a required item remains unobserved after sufficient
accumulated evidence, which is intended to reduce false positives from momentary
occlusions or detection noise.

\section{OSHA Violation Coverage}

Detected violations are mapped to specific OSHA standards, summarised in
Table~\ref{tab:osha}. Violations are scored by severity
(\texttt{LOW}--\texttt{CRITICAL}) and frequency; workers with repeated
violations are surfaced in the per-shift safety report.

\begin{table}[H]
\centering
\small
\caption{OSHA violation coverage by detection modality.}
\label{tab:osha}
\resizebox{\columnwidth}{!}{
\begin{tabular}{lll}
\toprule
\textbf{Standard} & \textbf{Violation} & \textbf{Source} \\
\midrule
1926.501 & Fall protection      & Wall cam \\
1926.503 & Training compliance  & Worker DB \\
1926.451 & Scaffolding          & Wall cam \\
1926.102 & Eye/face PPE         & Wall cam \\
1910.212 & Machine guarding     & Wall cam \\
---      & MSD / posture risk   & POV \\
---      & Ladder misuse        & Wall cam \\
---      & Respiratory PPE      & Wall cam \\
\bottomrule
\end{tabular}
}
\end{table}

\section{System Implementation}

The backend is a FastAPI application with async job processing, SQLite
persistence (via SQLAlchemy), and four core tables: \texttt{sites},
\texttt{workers}, \texttt{shifts}, and \texttt{safety\_events}. All models
(YOLO, SAM~3, Qwen3-VL) are pre-warmed on startup to avoid cold-start
memory errors. The wall-cam pipeline processes every third frame; the
posture pipeline runs concurrently in a thread pool. Safety events are
deduplicated per (worker, violation type) per shift. The system was
developed and validated on 4$\times$ NVIDIA RTX PRO 6000 Blackwell GPUs.

\section{Data and Datasets}

YOLO11 was fine-tuned on a composite PPE dataset assembled from Ultralytics
Construction-PPE (11 classes), Roboflow construction site safety, SH17,
and two additional Kaggle corpora. Posture threshold tuning used the CWPV
dataset~\cite{cwpv2024} (POV footage of construction workers annotated for
musculoskeletal posture) and the \"{O}nal \& Dand{\i}l behavioural
video dataset~\cite{onal2024} (691 clips, 8 behaviour classes). Validation
used long-form POV footage (15--20+ minutes per clip) from masonry,
preparation, and transit scenarios; compliant and non-compliant clips were
exported for threshold tuning and human review.

\section{Results}
\label{sec:results}

\begin{figure}[H]
\centering
\begin{subfigure}[t]{0.48\columnwidth}
    \includegraphics[width=\linewidth]{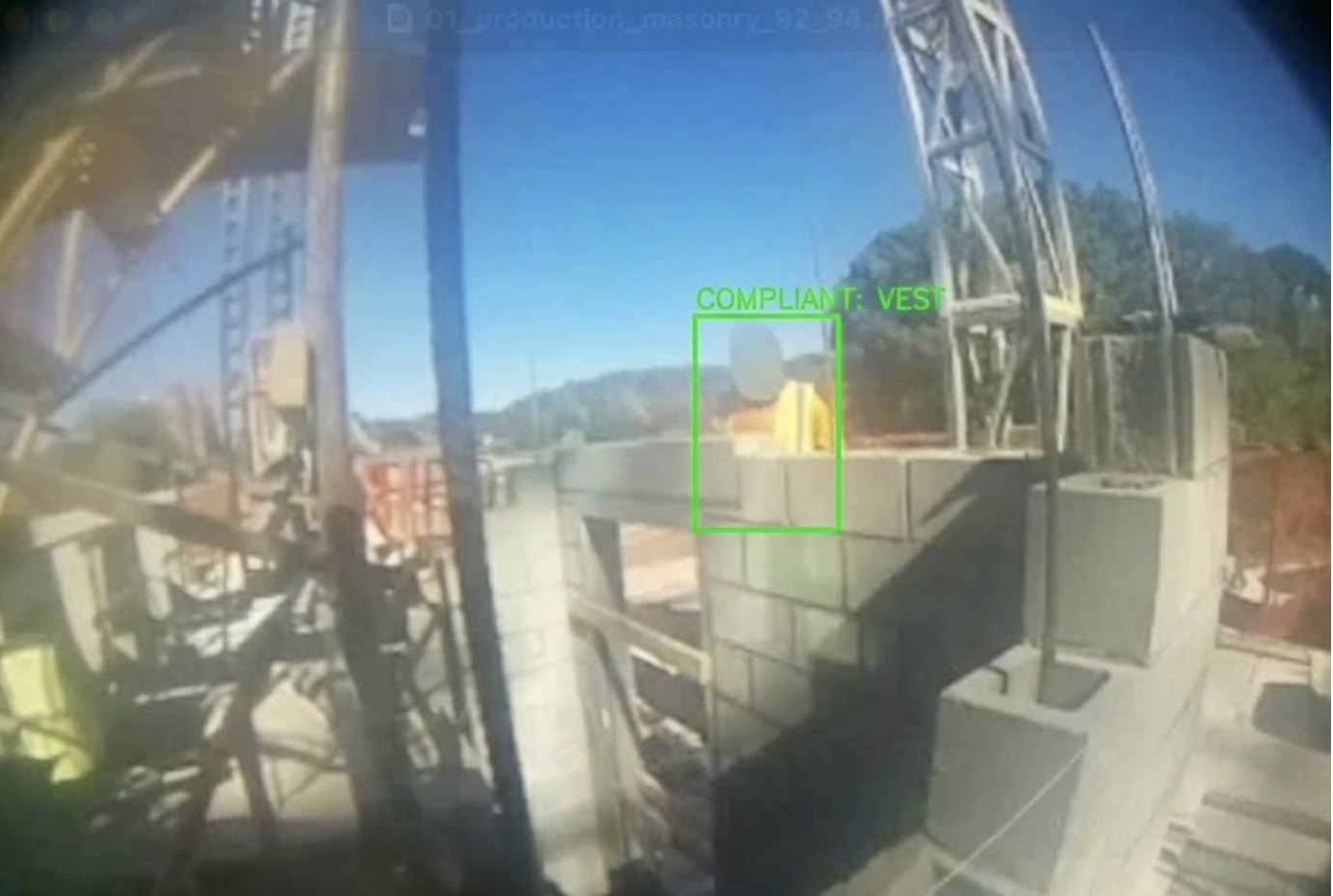}
    \caption{Compliant: safety vest detected.}
\end{subfigure}
\hfill
\begin{subfigure}[t]{0.48\columnwidth}
    \includegraphics[width=\linewidth]{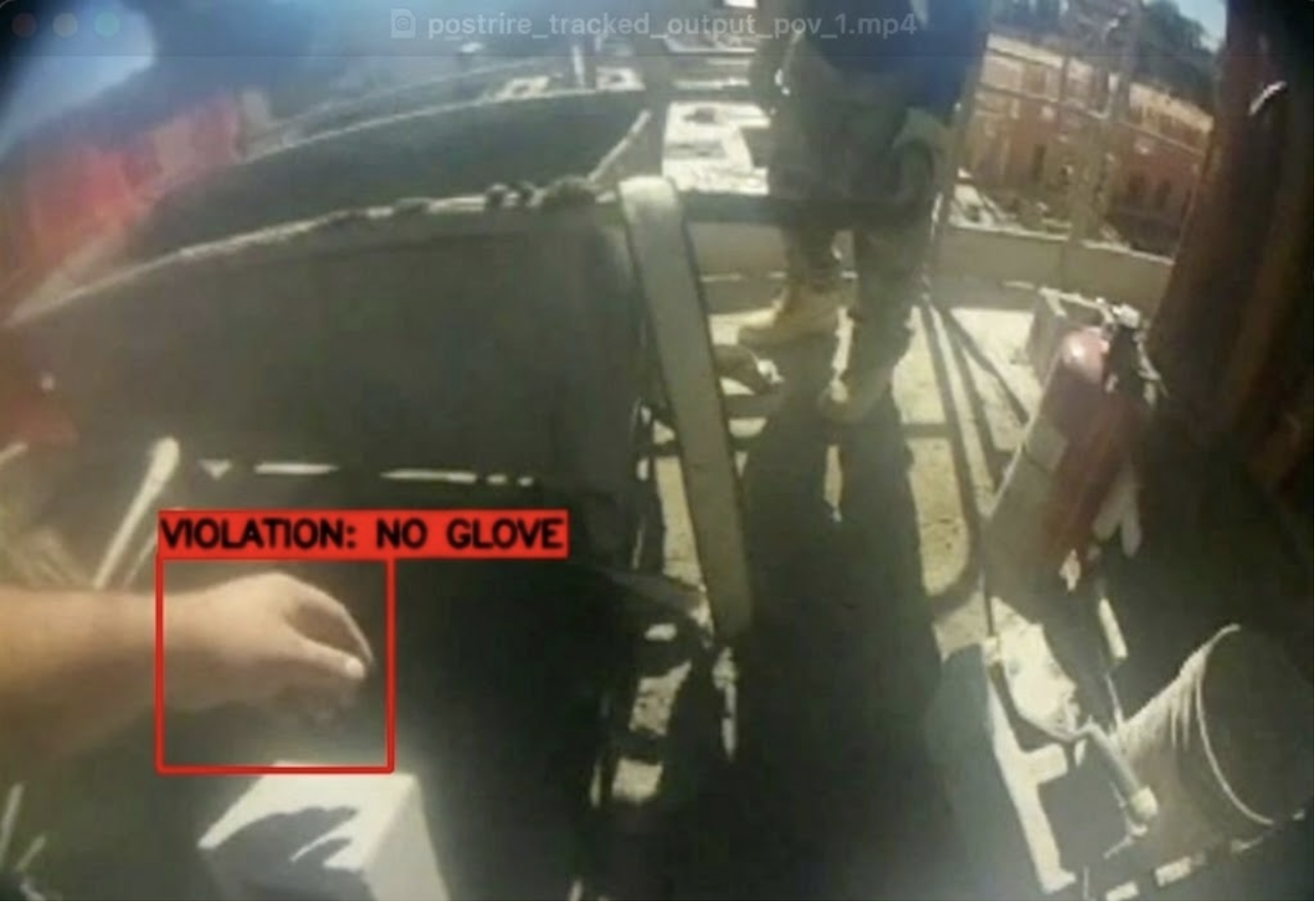}
    \caption{Violation: no glove flagged.}
\end{subfigure}
\caption{PPE detection results. (a)~Worker building a concrete block wall; 
system identifies the high-visibility vest and labels the worker as compliant. 
(b)~POV frame; system flags an exposed hand with a red bounding box for 
missing hand protection.}
\label{fig:ppe}
\end{figure}

\begin{figure}[H]
\centering
\begin{subfigure}[t]{0.48\columnwidth}
    \includegraphics[width=\linewidth]{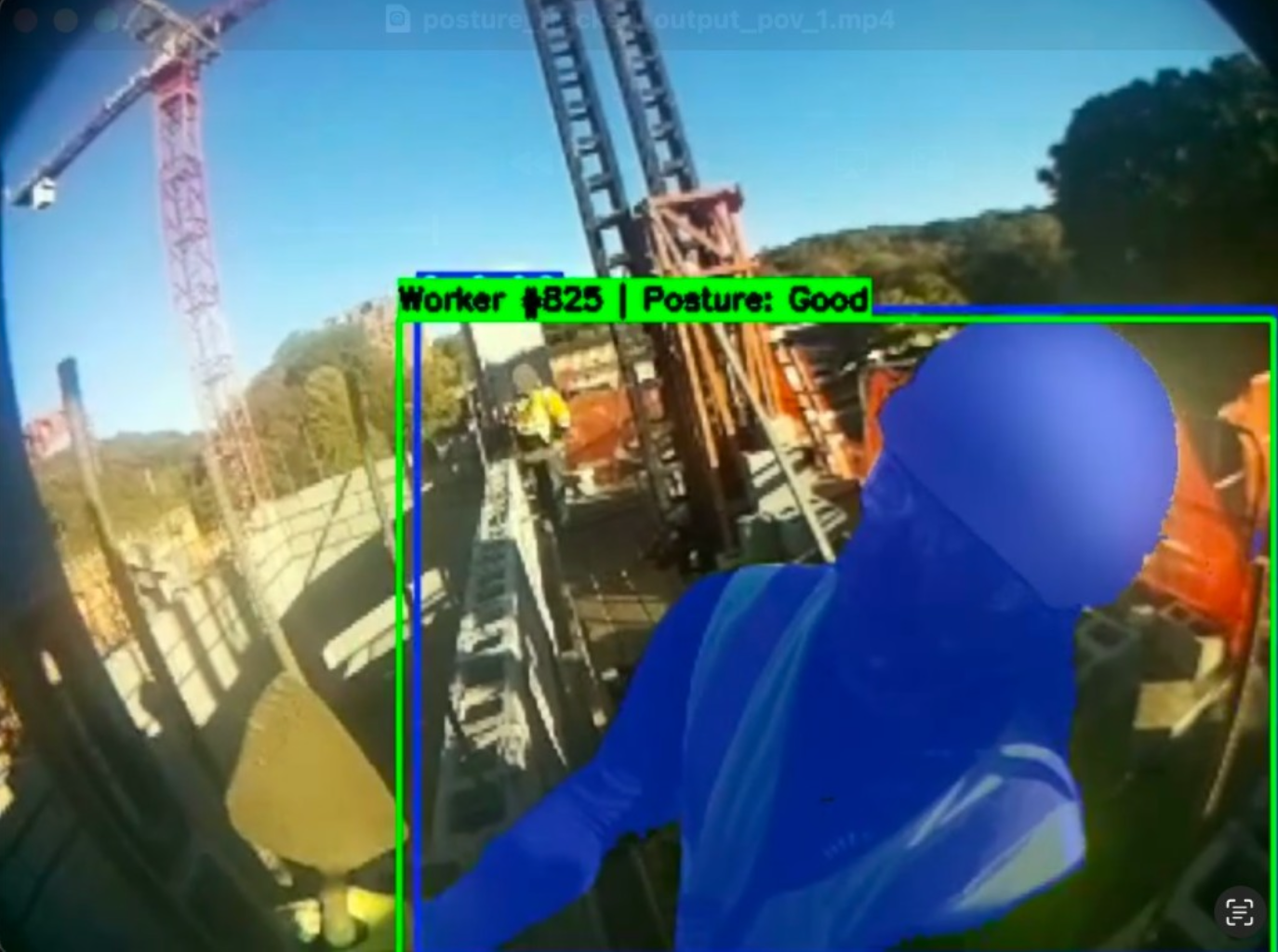}
    \caption{Worker \#825: Posture Good.}
\end{subfigure}
\hfill
\begin{subfigure}[t]{0.48\columnwidth}
    \includegraphics[width=\linewidth]{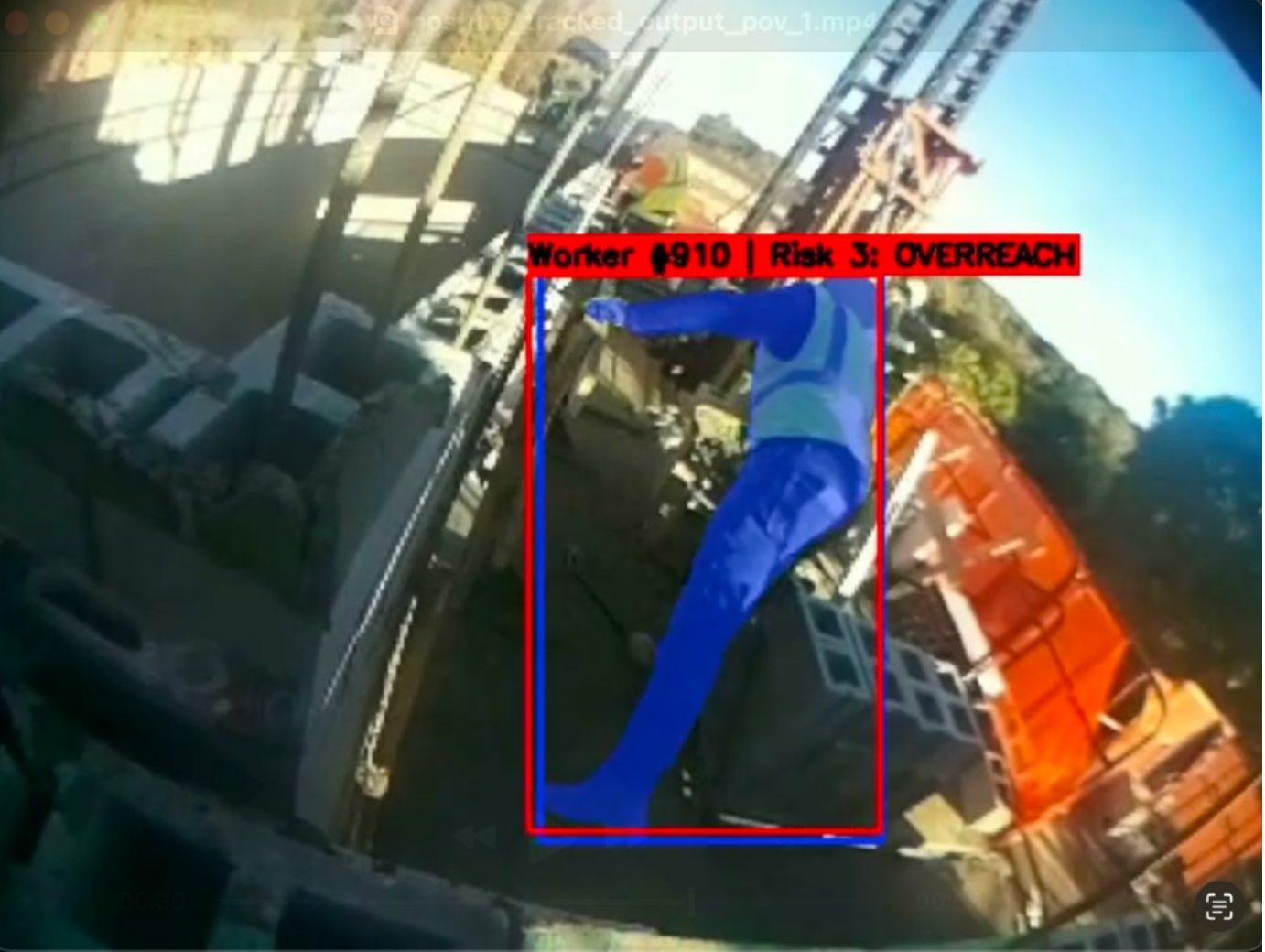}
    \caption{Worker \#910: Risk 3 --- Overreach.}
\end{subfigure}
\caption{Posture analysis results. (a)~Worker segmented with blue mask and 
green bounding box; pose and PPE assessed as compliant. (b)~Worker segmented 
with red box; elevated arm classified as \textsc{Overreach} at risk 
level~3 for musculoskeletal strain.}
\label{fig:posture}
\end{figure}

The pipeline was deployed end-to-end and exercised against multiple hours 
of construction site footage across both POV body-worn and fixed wall-mounted 
camera modalities under masonry, preparation, and transit task scenarios. 
Figures~\ref{fig:ppe} and~\ref{fig:posture} present representative outputs. 
Quantitative evaluation is deferred to future work pending re-provisioned 
GPU compute; the evaluation harness required to produce it is released 
alongside the codebase. All observations reported below are qualitative
and based on the development corpus.

\paragraph{PPE detection.}
On the development corpus, the Stage~1 detection layer identified hard hats
and high-visibility safety vests on wall-cam footage and recovered glove
presence on POV footage where the camera-wearer's hands occupied a substantial
portion of the frame. Figure~\ref{fig:ppe} shows a compliant detection
(safety vest identified, green label) alongside a violation (missing glove
flagged with red bounding box). The wall-cam pipeline resolves violations
from site-wide views, while POV cameras provide hand-level evidence for
glove compliance. Detection quality appeared to degrade under low-illumination
conditions; variable lighting remains an open limitation at the Stage~1
layer that warrants quantitative characterisation.

\paragraph{Posture and ergonomic analysis.}
On the development corpus, the REBA-inspired scoring distinguished routine
construction movement from visually unambiguous risk postures.
Figure~\ref{fig:posture} contrasts a compliant posture assessment with an
overreach violation. Arm-elevation events appeared recoverable from
pose-keypoint outputs on the inspected clips; the keypoint confidence gate
at $\tau_{\text{kp}} = 0.65$ appeared to suppress spurious detections from
partially occluded poses without visibly suppressing risk events on the
clips we inspected. Systematic measurement of false positive and false
negative rates is required to confirm these qualitative observations.

\paragraph{Three-pass adversarial VLM verification.}
The transition from single-pass VLM prompting to the three-pass adversarial
chain-of-thought protocol described in the Methodology produced the largest
single quality improvement observed during development. To estimate the
magnitude of this effect prior to controlled evaluation, we conducted an
informal review across the full 12-video Ironsite development corpus
(approximately 30~minutes per video, multiple violation instances per
video). Each video was processed by both the single-pass baseline and the
full three-pass protocol using the same Qwen3-VL-8B base model and an
equivalent total token budget. The three authors then independently
reviewed every violation flag emitted by each configuration and classified
it as a true or false positive by consensus, treating the three-author
consensus as local ground truth. Across all 12 videos, the three-pass
protocol produced approximately 12\% higher precision than the single-pass
baseline, with the largest gains on hallucination-prone categories where
violations are difficult to confirm from a single frame (gloves, eye
protection). We emphasise that this is an author-adjudicated evaluation
rather than a controlled study with blind annotation, and the same
authors who developed the system served as ground-truth raters; a
defensible measurement requires independent annotators on a held-out test
set, which the released evaluation harness is designed to produce. Early
single-pass runs exhibited confident emission of plausible but visually
unsupported violations in these categories; introducing the
generator/discriminator persona separation, the asymmetric reconciliation
rules (Section~\ref{sec:prompt-design}), and the explicit confidence-band
calibration appeared to reduce this failure mode to a level the development
team judged acceptable for end-of-shift reporting. A controlled ablation
isolating the contribution of each protocol component (persona separation
vs. confidence bands vs. asymmetric reconciliation) is implemented in the
released harness but not reported here.

\subsection{Limitations and Future Work}
Several limitations are noted, each suggesting a concrete direction for 
future work. The 12\% precision improvement reported in Section~\ref{sec:results}
was obtained through informal author-adjudication on the 12-video Ironsite
development corpus; the three authors who built the system also served as
ground-truth adjudicators, and no held-out test set or blind annotation
protocol was used. A controlled study with independent annotators,
inter-annotator agreement measurement, and a held-out test set is the
highest-priority subsequent activity, and the released evaluation harness
is designed to support it. Worker re-identification has not been
stress-tested in densely populated footage; the appearance-embedding
infrastructure is in place, but its accuracy at scale is unmeasured.
Variable lighting causes intermittent PPE false positives at the Stage~1
layer on the development corpus, suggesting domain-adaptive fine-tuning as
a natural extension. The posture threshold boundary between safe and
borderline ergonomic risk remains context-dependent and may require
per-site calibration. Finally, POV cameras frequently experience mid-task
view obstruction; multi-camera fusion across simultaneous POV and
wall-mounted streams is a promising mitigation the present architecture
supports but does not yet exploit.
\section{Discussion}

The architectural decisions and qualitative observations reported here suggest
that end-of-shift passive monitoring may be a viable alternative to real-time
surveillance for construction safety compliance, pending quantitative
validation. Several design decisions that emerged during development warrant
discussion.

VLM hallucination was the most prominent technical obstacle encountered.
Early single-pass runs produced confident violation reports that were
visually unsupported, a failure mode consistent with findings in the
broader VLM reliability literature~\cite{Baietal2024}. The adversarial
multi-pass architecture --- in which two structurally independent observer
personas assess the same scene before a reconciliation pass --- was the
intervention that appeared, qualitatively, to most reduce observed
hallucination during development. The asymmetric reconciliation rules
(Section~\ref{sec:prompt-design}), which treat behavioural violations as
observer-strong and spatial PPE violations as machine-strong, reflect our
working hypotheses about the complementary detection characteristics of
human observation and automated detection respectively. These hypotheses
remain to be tested.

PPE-to-worker association proved nontrivial under worker clustering. A
detected hard hat occupying the airspace between two adjacent workers
does not automatically belong to the worker whose bounding box overlaps
it most; centre-point containment logic combined with temporal
accumulation across frames was useful for more reliable association on
the clips we inspected. Ergonomic threshold selection required extensive
iteration: the distinction between routine construction movement and
genuinely hazardous posture is context-dependent, and a fixed threshold
--- such as a $45^\circ$ trunk flexion limit --- can misclassify momentary
tool pickups and sustained awkward postures alike without additional
temporal context. Joint GPU residency for three large models demanded careful
orchestration; pre-warming on startup, sequential inference staging, and
60-second video chunking were necessary to prevent out-of-memory failures
on the development hardware.

\section{Conclusion}

A passive, end-of-shift construction safety monitoring pipeline has been
described, processing video from POV body-worn and fixed wall-mounted
cameras through a unified three-stage architecture: fine-tuned YOLO
detection, SAM~3 segmentation refinement, and adversarial VLM
verification. The system maps detected violations to specific OSHA
standards, performs REBA-inspired ergonomic risk scoring from pose
keypoints, and produces per-worker safety reports with timestamped
frame-level evidence. The central methodological element is a three-pass
adversarial chain-of-thought protocol in which structural message
isolation is used to enforce observational independence between passes,
and asymmetric reconciliation rules encode explicit working assumptions
about the relative reliability of human observation versus automated
detection for different violation categories. Qualitative observations on
construction site footage are consistent with the feasibility of the
approach; quantitative evaluation against the released harness is
deferred to future work pending re-provisioned GPU compute.

Future work should prioritise: controlled quantitative ablation of the
three-pass VLM protocol against single-pass and two-pass baselines;
stress-testing of the appearance-embedding re-identification module in
densely populated multi-worker footage; domain-adaptive fine-tuning of
the Stage~1 detector on low-illumination construction footage; and
integration with EHS reporting platforms such as Procore and Autodesk
Construction Cloud. Longer term, real-time ingestion via RTSP/WebRTC
and multi-camera fusion across simultaneous POV and wall-mounted streams
represent natural extensions that the present architecture supports but
does not yet exploit.

\appendix
\onecolumn
\section{Supplementary Material: VLM System Prompts}
\label{app:prompts}

The full system prompts used in the three-pass adversarial chain-of-thought
protocol are reproduced verbatim below. The prompts are referenced from
Section~\ref{sec:prompt-design}.

\begin{figure}[H]
\small
\begin{mdframed}
\textbf{Pass 1 System Prompt (Jamie Reyes --- Generator)}\\[4pt]
\textit{You are Jamie Reyes, a field safety inspector with 6 years of on-site
construction experience. You are conducting an initial walkthrough review of this
site camera footage and filing a written inspection report.}

\textit{Your report will be reviewed and audited by Marcus Chen --- Chief Safety
Officer with 24 years of experience. He will be comparing your findings against
machine-detection data and annotated frames from an AI system. If you miss
something obvious or write vague non-observations, he will catch it and it will
reflect on your competence.}

\textit{Be thorough, honest, and specific. Name what you see. If something concerns
you, write it down clearly --- do not hedge into uselessness. If something looks
fine, say so and say why. Note approximate timestamps where helpful. Write in plain
English, not JSON. This is your inspection report, not a final verdict --- Marcus
will have the final say.}
\end{mdframed}
\caption{Pass~1 system prompt. The generator receives raw video only; no machine
detection data or prior notes are present in the message array.}
\label{fig:prompt1}
\end{figure}

\begin{figure}[H]
\small
\begin{mdframed}
\textbf{Pass 2 System Prompt (Marcus Chen --- Discriminator)}\\[4pt]
\textit{You are Marcus Chen. You have 24 years as a senior construction safety
manager and Chief Safety Officer. You are conducting your own independent review
of a site camera video.}

\textit{You have NOT seen any other inspector's report. You are watching the raw
video yourself, and you also have the machine-annotated frames from the AI detection
system (YOLO + SAM) showing bounding boxes and segmentation masks over flagged
moments.}

\textit{Watch the raw video with your own eyes first. Then cross-reference with the
annotated frames. Your job: write your independent professional assessment.}
\begin{itemize}[nosep, leftmargin=*]
\item \textit{What do you observe in the raw video? Any workers, PPE, hazards,
unsafe behaviour?}
\item \textit{For each machine-flagged detection: do you agree based on what
you see?}
\item \textit{Did the machine miss anything you can see clearly in the raw footage?}
\item \textit{Note detections that look like noise or false positives.}
\end{itemize}
\textit{Write in plain English. Be direct. Do NOT produce JSON --- a third pass
will do that.}
\end{mdframed}
\caption{Pass~2 system prompt. The discriminator receives raw video, annotated
frames, and YOLO confidence data, but no Pass~1 output.}
\label{fig:prompt2}
\end{figure}

\begin{figure}[H]
\small
\begin{mdframed}
\textbf{Pass 3 Reconciliation Rules (verbatim from user template)}\\[4pt]
\textit{You have three sources of evidence. Reconcile them and produce the final
safety assessment.}

\textbf{SOURCE 1} --- Jamie's field inspection report (raw video, no machine data)\\
\textbf{SOURCE 2} --- Your own assessment of the annotated frames\\
\textbf{SOURCE 3} --- YOLO + SAM detection data\\[4pt]

\textbf{RECONCILIATION RULES:}
\begin{itemize}[nosep, leftmargin=*]
\item \textit{AGREEMENT (Jamie + you both flagged it, YOLO confirms) $\rightarrow$
flag it, high confidence.}
\item \textit{SPLIT (Jamie flagged, you did not) $\rightarrow$ DO NOT automatically
dismiss Jamie. Jamie watched the raw video --- he catches motion, behaviour, and
momentary events that annotated frames may not show$\ldots$ Require a specific reason
to overrule Jamie, not just absence of machine detection.}
\item \textit{MACHINE ONLY (YOLO flagged, neither observer noted it) $\rightarrow$
conf $\geq$ 0.70 = flag with note; conf $<$ 0.40 = discard.}
\item \textit{OBSERVER ONLY (Jamie or you flagged it clearly, YOLO missed it)
$\rightarrow$ flag it. Machines miss things constantly. A clear human observation
of a violation is sufficient.}
\item \textit{$\Rightarrow$ DO NOT write ``no violations'' if Jamie's notes contain
specific flagged observations. Engage with each one explicitly.}
\end{itemize}
\end{mdframed}
\caption{Pass~3 reconciliation rules (user template, verbatim). Pass~3
omits the raw video and operates on committed textual evidence plus annotated frames.}
\label{fig:prompt3}
\end{figure}

\end{document}